\documentclass{article}

\PassOptionsToPackage{numbers, compress}{natbib}

\usepackage[accepted]{icml_fmwild}

\usepackage[utf8]{inputenc} %
\usepackage[T1]{fontenc}    %
\usepackage{amsmath}
\usepackage{amssymb}
\usepackage{mathtools}
\usepackage{amsthm}
\usepackage{amsfonts}       %
\usepackage{url}            %
\usepackage{booktabs}       %
\usepackage{nicefrac}       %
\usepackage{microtype}      %
\usepackage{xcolor}         %
\usepackage{bm}
\usepackage{diagbox}
\usepackage{oplotsymbl}
\usepackage{pgfplots}
\usepackage{wrapfig}
\usetikzlibrary{patterns}

\usepackage{caption}
\DeclareMathOperator*{\argmax}{arg\,max}

\newif\ifdraft
\draftfalse

\usepackage{xparse}
\usepackage{xspace}
\usepackage{algorithm}%
\usepackage{fixltx2e}
\usepackage{tikz}
\usepackage{float}
\usepackage{multirow}
\usepackage{multicol}
\usepackage{colortbl}
\usepackage{array}
\newcommand{\PreserveBackslash}[1]{\let\temp=\\#1\let\\=\temp}
\newcolumntype{C}[1]{>{\PreserveBackslash\centering}p{#1}}
\newcolumntype{R}[1]{>{\PreserveBackslash\raggedleft}p{#1}}
\newcolumntype{L}[1]{>{\PreserveBackslash\raggedright}p{#1}}

\usepackage{microtype}
\usepackage{graphicx}
\usepackage{mathtools}
\usepackage{float}
\usepackage{booktabs} %
\usepackage[shortlabels]{enumitem}

\usepackage{amssymb}%
\usepackage{pifont}%

\usepackage{bbold}

\usepackage{hyperref,amssymb,enumitem}

\setlist[itemize]{leftmargin=*}
\setlist[enumerate]{leftmargin=*}

\newcommand*{\rej}{{\ooalign{\lower.3ex\hbox{$\sqcup$}\cr\raise.4ex\hbox{$\sqcap$}}}}

\usepackage{stackengine}

\usepackage{xcolor}

\renewcommand{\algorithmicrequire}{\textbf{Input: }}
\renewcommand{\algorithmicensure}{\textbf{Output: }}

\newcommand{\ie}{\textit{i.e.,}\@\xspace}
\newcommand{\eg}{\textit{e.g.,}\@\xspace}

\graphicspath{{images/}}

\usepackage{booktabs,arydshln}
\makeatletter
\def\adl@drawiv#1#2#3{%
        \hskip.5\tabcolsep
        \xleaders#3{#2.5\@tempdimb #1{1}#2.5\@tempdimb}%
                #2\z@ plus1fil minus1fil\relax
        \hskip.5\tabcolsep}
\newcommand{\cdashlinelr}[1]{%
  \noalign{\vskip\aboverulesep
           \global\let\@dashdrawstore\adl@draw
           \global\let\adl@draw\adl@drawiv}
  \cdashline{#1}
  \noalign{\global\let\adl@draw\@dashdrawstore
           \vskip\belowrulesep}}
\makeatother

\newcommand{\nlp}[1]{}

\newcolumntype{x}[1]{>{\centering\arraybackslash\hspace{0pt}}p{#1}}

\ifdraft

\usepackage{xargs}                      %

\else

\fi

\usepackage[colorinlistoftodos,prependcaption,textsize=tiny]{todonotes}

\newcommand\xintu{0.5pt}

\newcommand{\EmbedAttack}{\textsl{EmbedAttack}\@\xspace}
\newcommand{\PGDAttacks}{\textsl{PGDAttacks}\@\xspace}

\newcommand{\SegPGD}{\textsl{SegPGD}\@\xspace}
\newcommand{\PGD}{\textsl{PGD}\@\xspace}
\newcommand{\DepthPGD}{\textsl{DepthPGD}\@\xspace}

\usepackage{amsmath,amsfonts,bm}

\def\eqref#1{equation~\ref{#1}}

\def\1{\bm{1}}

\DeclareMathAlphabet{\mathsfit}{\encodingdefault}{\sfdefault}{m}{sl}
\SetMathAlphabet{\mathsfit}{bold}{\encodingdefault}{\sfdefault}{bx}{n}

\usepackage{svg}
\usepackage{graphicx}
\usepackage{subcaption}
\usepackage{hyperref}       %
\usepackage[capitalize,noabbrev]{cleveref}

\icmltitlerunning{Benchmarking Robust Self-Supervised Learning Across Diverse Downstream Tasks}

\begin{document}

\twocolumn[
\icmltitle{Benchmarking Robust Self-Supervised Learning\\Across Diverse Downstream Tasks}

\icmlsetsymbol{equal}{*}

\begin{icmlauthorlist}
\icmlauthor{Antoni Kowalczuk}{equal,cispa}
\icmlauthor{Jan Dubiński}{equal,wut,ideas}
\icmlauthor{Atiyeh Ashari Ghomi}{equal,L6}
\icmlauthor{Yi Sui}{L6}
\icmlauthor{George Stein}{L6}
\icmlauthor{Jiapeng Wu}{L6}
\icmlauthor{Jesse C. Cresswell}{L6}
\icmlauthor{Franziska Boenisch}{cispa}
\icmlauthor{Adam Dziedzic}{cispa}
\end{icmlauthorlist}

\icmlaffiliation{cispa}{CISPA Helmholtz Center for Information Security}
\icmlaffiliation{wut}{Warsaw University of Technology}
\icmlaffiliation{ideas}{IDEAS NCBR}
\icmlaffiliation{L6}{Layer 6 AI, Toronto, Canada}

\icmlcorrespondingauthor{Adam Dziedzic}{adam.dziedzic@sprintml.com}

\icmlkeywords{}

\vskip 0.3in
]

\printAffiliationsAndNotice{\icmlEqualContribution} %

\begin{abstract}

Large-scale vision models have become integral in many applications due to their unprecedented performance and versatility across downstream tasks. However, the robustness of these foundation models has primarily been explored for a single task, namely image classification. The vulnerability of other common vision tasks, such as semantic segmentation and depth estimation, remains largely unknown. We present a comprehensive empirical evaluation of the adversarial robustness of self-supervised vision encoders across multiple downstream tasks. Our attacks operate in the encoder embedding space and at the downstream task output level. In both cases, current state-of-the-art adversarial fine-tuning techniques tested only for classification significantly degrade clean and robust performance on other tasks. Since the purpose of a foundation model is to cater to multiple applications at once, our findings reveal the need to enhance encoder robustness more broadly. %
Our code is available at \href{https://github.com/layer6ai-labs/ssl-robustness}{github.com/layer6ai-labs/ssl-robustness}.

\end{abstract}

\section{Introduction}

Foundation models trained through self-supervised learning (SSL) have become the backbone of many applications due to their versatility; one foundation model can be adapted to many downstream tasks with a small amount of data and training (or fine-tuning). Foundation models in the vision domain have even outperformed dedicated models on several tasks~\citep{caron2021dino,mae,oquab2024dinov2}.
Despite their broad utility, the adversarial robustness of these models has only been explored for classification tasks with linear probing~\citep{NaseerPurifier2020CVPR,jiang2020robust,fan2021does,zhang2022decoupled,luo2023rethinking} while other common downstream tasks, such as semantic segmentation~\citep{long2017segmentation} and depth estimation~\citep{godard2019depthestimation}, remain unexplored.  Recently, ~\citet{li2023adversarialnotbugs} showed that non-robust features extracted from adversarial examples for supervised models (and useful for classification) become largely useless when transferred to self-supervised learning paradigms. They advocated for a cross-paradigm examination of robustness, yet focused their analysis solely on classification. A major outstanding question is whether adversarial robustness transfers across downstream tasks.

\begin{figure}[t]
    \centering
    \includegraphics[width=1.0\linewidth, trim={26 10 0 0}, clip]{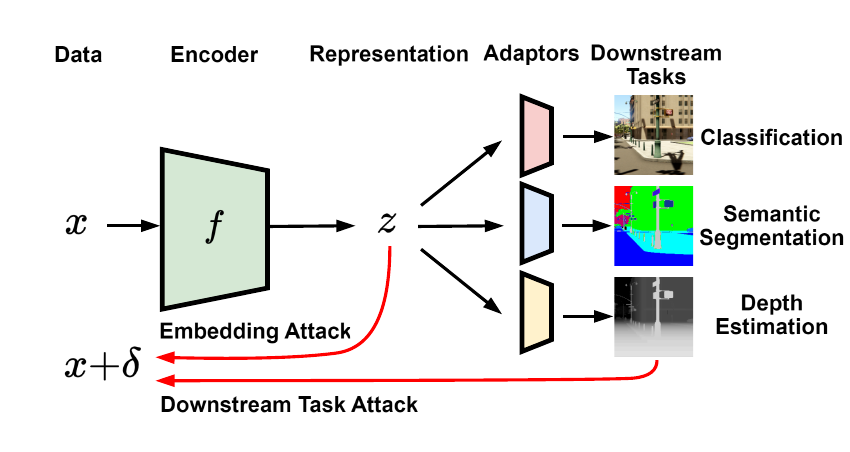}
    \vspace{-18pt}
    \caption{\textbf{ Adversarial Attacks for SSL.} An SSL encoder is applied to downstream tasks through adaptors. Adversarial attacks can attack the downstream labels, or the embedding space directly.}
    \label{fig:ssl}
    \vspace{-2em}
\end{figure}

We present an in-depth empirical evaluation of the adversarial robustness of self-supervised vision encoders~\citep{simsiam,caron2021dino} for downstream tasks beyond classification. We use attacks that operate in the encoder's embedding space (\EmbedAttack) and those that leverage direct access to the downstream task outputs (\PGDAttacks), \eg \PGD for classification~\citep{madry2018towards} or \SegPGD for semantic segmentation~\citep{gu2022segpgd}. Our main observation is that the state-of-the-art adversarial full fine-tuning of encoders~\citep{zhang2022decoupled}: (1) substantially lowers clean performance, (2) increases robustness only against the \EmbedAttack, and (3) remains ineffective in improving robustness against the task-specific \PGDAttacks. We observe only a slight improvement against the \PGDAttacks for classification when the adversarial fine-tuning dataset and downstream dataset come from the same distribution. This indicates a need to rethink what it means for a foundation model to be robust. Finally, we offer potential approaches to bolster the cross-task robustness of SSL encoders.%

\section{Background and Related Work}
\label{attack_framework}

\textbf{Self-Supervised Learning.} 
SSL aims to extract a representation of data which is useful for downstream tasks specified at test-time \citep{balestriero2023cookbook}. In many 
frameworks, an input $x$ is first modified by two semantic-preserving augmentations yielding $x_1$ and $x_2$, which are subsequently passed to an encoder $f$. The training objective aligns the output representations by minimizing a distance metric $d$ (\eg Euclidean distance) as $L(f, x) = d(f(x_1),f(x_2))$ \citep{chen2020simple}. Once trained, the encoder's representations are then used for various downstream tasks, such as \textit{classification}, \textit{semantic segmentation}, or \textit{depth estimation} by fine-tuning adaptor networks. In this work, we focus on a state-of-the-art SSL framework, \textbf{DINO}~\citep{caron2021dino}. DINO utilizes two encoder networks, the \textit{teacher} $f_t$, and \textit{student} $f_s$. The student network is optimized to minimize the cross-entropy between $f_s(x_1)$ and the soft labels $f_t(x_2)$, as a form of knowledge distillation \citep{hinton2015distilling}. To prevent collapse, the gradients are only passed through $f_s$. Parameters of $f_t$ are updated using the moving average of the student's parameters. \textbf{DINOv2}~\citep{oquab2024dinov2} improves over DINO in terms of scale and efficiency of training, rather than proposing a new SSL method. \citet{oquab2024dinov2} showed substantial improvements on dense (pixel-wise) downstream tasks like semantic segmentation and depth estimation compared to DINO encoders.

\textbf{Adversarial Robustness in SSL.} In this work, we focus on the state-of-the-art \textbf{Decoupled Adversarial Contrastive Learning (DeACL)} framework by~\cite{zhang2022decoupled} to obtain robust SSL encoders. For an overview on other methods for robust SSL and a thorough discussion on the advantages of DeACL, see \Cref{app:related_work_robustness}.
DeACL fine-tunes existing encoders for increased robustness using knowledge transfer from a pre-trained encoder to a robust one. The objectives for the distillation are to: (1) match the distilled encoder representations to those of the pre-trained encoder (high cosine similarity), and (2) bring the distilled encoder's representations of adversarial examples (\ie examples generated with the pre-trained encoder that maximize the distance to their original samples) close to their clean counterparts. By decoupling the encoder pre-training from increasing its robustness, DeACL provides high computational efficiency in comparison to prior methods and obtains state-of-the-art robust performance.

\textbf{Downstream Tasks.}
To evaluate the quality of representations learned by SSL methods, we consider three common downstream tasks. 
\textbf{(1) Linear Classification} assesses the quality of the learned representations by training a downstream classifier and measuring classification performance.
\textbf{(2)~Semantic Segmentation} is a common computer vision task that categorizes every pixel in an image into a class or object. While downstream-agnostic adversarial examples against SSL encoders can be used to fool segmentation models, \citet{gu2022segpgd} show with \SegPGD that tailoring the attack to the segmentation task is even more effective. \SegPGD aims at manipulating all pixel classifications of an image by introducing a weighted loss term between correctly classified and misclassified pixels.
\textbf{(3)~Depth Estimation} is another prevalent computer vision task aimed at estimating distances of objects in an image relative to the camera location, where each pixel is assigned a depth value.
Targeted adversarial attacks against depth estimation can lead to strong deviations between actual and predicted depth~\citep{wong2020targeted}. At the same time, they can also be leveraged for depth estimation-specific adversarial training to improve robustness~\citep{cheng2022adversarial}.

\section{Attack and Defense Methods}
We propose a framework to assess the robustness of foundation models at both the embedding level and for downstream tasks, as described in \Cref{attack_framework}. 
The goal of benchmarking the robustness of foundation models across diverse downstream tasks restricts our possible selection of encoder models. Specifically, the encoder must generate representations that are applicable to a variety of tasks beyond classification. In our preliminary experiments, we evaluated the performance of SimCLR \cite{chen2020simple}, SimSiam \cite{simsiam}, and DINO encoders. We observed that the representations produced by SimCLR and SimSiam were insufficient to achieve high-quality downstream segmentation or depth estimation. 
For that reason, we use the foundation models DINO and DINOv2 as examples, and train a linear adaptor for each downstream task. For the embedding attack, we target the model at the representation layer. For downstream attacks, we evaluate three different tasks: classification, semantic segmentation, and depth estimation. Each attack is detailed in the following sections.
\subsection{Embedding-level Attack} The \EmbedAttack operates directly on the underlying encoder's embeddings~\cite{kim2020adversarial,jiang2020robust,fan2021does,luo2023rethinking}. The objective behind the approach is to make imperceptibly small modifications to an input image such that the resulting representation from the SSL encoder is changed substantially. 
More concretely, for a clean input image $x$, we find its adversarial perturbation $x_{\text{adv}}=x+\delta$ such that $\Vert \delta\Vert_{\infty} < \varepsilon$, where $\varepsilon$ is the maximum allowed input distortion measured in the $\ell_{\infty}$-norm. Given an encoder $f$, the objective is to find $x_{\text{adv}}$ such that the $\ell_2$ distance between the representations from the original $f(x)$ and adversarial images $f(x_{\text{adv}})$ is maximized: $\argmax_{x_{\text{adv}}} \Vert f(x) - f(x_{\text{adv}})\Vert_2$. 
For sparse downstream tasks (classification) we target the CLS token embedding, while for dense tasks (semantic segmentation and depth estimation) we target patch embeddings. 
We leverage the projected gradient descent (\PGD) attack~\citep{madry2018towards} with the objective defined in representation space to find adversarial examples $x_{\text{adv}}$. We set the maximum perturbation to $\varepsilon = 8/255$, start with $x_{\text{adv}}$ initialized from $x$ with uniform noise added (defined as $\mathcal{U}(-\varepsilon,\varepsilon)$), and perform 20 steps of \PGD with step size $2/255$. To ensure that the distance from the original image $x$ is within the $\varepsilon$-ball, we clip the perturbation to $[-\epsilon,\epsilon]$ at every step of \PGD. 

\subsection{Downstream Attacks}
\textbf{Classification.} 
For the standard classification tasks, we use the \PGD attack~\cite{madry2018towards} with settings similar to those above: $\varepsilon = 8/255$, 20 steps with step size $2/255$, and initialization from randomly perturbed images. The target is to maximize cross-entropy loss for the perturbed images.

\textbf{Semantic Segmentation.} 
To attack semantic segmentation we leverage the \SegPGD attack \cite{gu2022segpgd} which calculates a weighted average of the loss over correctly and incorrectly classified pixels,
\begin{equation}
    L(f_{\text{seg}}(x^t_{\text{adv}}), y) = \frac{1 - \lambda_t}{HW} \sum_{j \in P_T} L_j +  \frac{\lambda_t}{HW} \sum_{k \in P_F} L_k.
\end{equation}
Here $L_j$ represents the cross-entropy loss, $\lambda_t$ is a hyperparameter, $H$ and $W$ denote the height and width of the image, while $P^T$ and $P^F$ are the sets of correctly and incorrectly classified pixels respectively. \PGD is used to find adversarial examples with this loss, and we use similar settings as mentioned previously. The weight $\lambda_t$ starts from zero and increases linearly each iteration. The main insight behind the \SegPGD attack is to fool correctly classified pixels in the first attack iterations and then treat the correct and incorrect pixel classifications roughly equally in the later iterations. As a result, the \SegPGD attack can achieve similar attack effectiveness as \PGD but with substantially fewer iterations.

\textbf{Depth Estimation.} Similarly to semantic segmentation, we compute the average loss per pixel and then apply a \PGD attack targeting this loss, referred to as \DepthPGD. The loss terms used for depth estimation and its attack are akin to those in \citet{oquab2024dinov2}, incorporating the multi-scale gradient matching loss \cite{MegaDepthLi18} and pixel-wise depth loss \cite{AdaBins}. For more details, refer to \Cref{sec:hyperparameters_app}.

\subsection{DeACL Defense}
We combat the above attacks using the state-of-the-art method of obtaining robust encoders, DeACL~\citep{zhang2022decoupled}.
We select DeACL for several compelling reasons. Firstly, unlike many other methods aimed at enhancing robustness, it does not rely on any specific downstream task and instead improves robustness at the representation layer in a self supervised manner. Besides, it is a state-of-the-art method with superior robustness compared to other techniques. Lastly, the proposed adversarial fine-tuning approach is significantly more computationally efficient compared to training models from scratch using traditional adversarial training methods. These advantages make DeACL feasible and practical, particularly given the substantial computational resources required to train state-of-the-art encoder models. 
We start from a pretrained encoder $f$ and create its robust version $f_{R}$ using fine-tuning with the following objective: 
\begin{equation}
    L(f_R, f) = d(f_R(x), f(x)) + \gamma d(f_R(x_{\text{adv}}), f_R(x))\text{.}
    \label{eq:deacl_obj}
\end{equation}
Here we set $d$ as the standard cosine similarity, and $x$ as the input image. \Cref{eq:deacl_obj} aims to preserve representation quality, and improve robustness against adversarial examples. $\gamma=2$ is a parameter used to balance the impact of each goal on the final objective function.

\section{Empirical Evaluation}

\subsection{Setup}
We present the results for encoders trained using the DINO and DINOv2 SSL frameworks, utilizing ViT~\citep{dosovitskiy2020image} backbones. The underlying encoders are either \textit{Standard}, \ie provided by the SSL frameworks, or \textit{DeACL}, further fine-tuned to enhance robustness. We present the hyperparameters that we use to train the linear layers for the various of downstream tasks. These hyperparameters are uniform across encoders and datasets, and vary only between different types of tasks, \ie classification, semantic segmentation, and depth estimation. Full insights are presented in \Cref{sec:hyperparameters_app}. 

\textbf{Classification.} We use a learning rate of 0.5, batch size 16, and train the linear classifiers for 5 epochs using the Adam~\citep{kingma2014adam} optimizer. As a train-time augmentation we use random horizontal flips.

\textbf{Semantic segmentation.} We follow the setup from the DINOv2 framework, and use a learning rate of 0.0001, batch size 16, weight decay 0.001, and train for 50 epochs using the AdamW~\citep{loshchilov2018decoupled} optimizer. For training as well as evaluation on non-uniformly sized images (\eg PASCAL VOC 2012) we utilize sliding window inference, \ie we divide the image into parts of uniform size, compute logits for all of the parts, and then combine them into one final logit map. Overlap between the parts is handled by averaging the logits in the overlap regions. We use random cropping, and random horizontal and vertical flips as training-time augmentations.

\textbf{Depth estimation.}
Since DINOv2 has achieved state-of-the-art performance in depth estimation, we adopt their settings. For training, we use their combination of gradient matching loss and pixel-wise depth loss. For the remaining hyperparameters, we use a learning rate of 0.0001, batch size 128, weight decay 0.01, and train for 20 epochs using AdamW. All hyperparameters are listed in \Cref{sec:depth-hyper}.

\subsection{Results}
\textbf{Classification.} 
We follow the widely used linear evaluation protocol~\citep{chen2020simple,simsiam}, where a linear classifier is trained on top of the frozen base SSL encoder, and test accuracy is used as a proxy for representation quality. We compare the classification accuracy after linear probing for the standard vision benchmarks: CIFAR10~\citep{krizhevsky2009learning}, CIFAR100~\citep{krizhevsky2009learning}, and STL10~\citep{coates2011analysis}. The evaluation is presented in \Cref{tab:classification}. Contrary to the results shown by~\citet{zhang2022decoupled}, we observe no improvement in robustness against tailored \PGD attacks (right column) for the encoder fine-tuned using DeACL, with the only exception being on the STL10 dataset. We argue that the discrepancy in our results and ones reported by \citet{zhang2022decoupled} stems from the underlying training sets of the fine-tuned encoder. \citet{zhang2022decoupled} utilized encoders trained on CIFAR10, then fine-tuned and evaluated them on CIFAR10 as well. In contrast, we focus on ImageNet-trained encoders, use ImageNet for fine-tuning, and evaluate them on various datasets including CIFAR10. We assume that the discrepancy between training, fine-tuning, and evaluation sets leads directly to the inefficacy of DeACL in obtaining robust encoders against stronger adversarial attacks than \EmbedAttack, like \PGD. This idea is supported by the improved adversarial accuracy against \PGD attacks on STL10 with the fine-tuned encoder, as it is a subset of ImageNet. We observe an increase (above random guessing) in accuracy compared to the standard encoder (see second last and the last row of \Cref{tab:classification}, rightmost column), from 0 to 0.23.

\renewcommand\xintu{1.0pt}
\addtolength{\tabcolsep}{-\xintu}
\begin{table}[h!]
    \tiny
    \centering
    \setlength{\tabcolsep}{3pt}
    \begin{tabular}{ccc ccc}
    \toprule 
    Dataset & SSL & Encoder & \textit{Clean} & \EmbedAttack & \PGD \\
    & Framework & Type & Accuracy$\uparrow$ & Accuracy$\uparrow$ & Accuracy$\uparrow$ \\[0.0cm]  \midrule
    CIFAR10 & DINO v2 ViT-S/14 & Standard & 0.94 & 0.01 & 0.00 \\
    CIFAR10 & DINO v2 ViT-B/14  & Standard & 0.98 & 0.04 & 0.00 \\ 
    CIFAR10 & DINO v1 ViT-B/16 & Standard & 0.94 & 0.01 & 0.00 \\ \hline
    CIFAR10 & DINO v1 ViT-B/16 & DeACL & 0.91 & 0.73 & 0.02 \\ \midrule
    
    CIFAR100 & DINO v2 ViT-S/14 & Standard & 0.82 & 0.00 & 0.00 \\
    CIFAR100 & DINO v2 ViT-B/14  & Standard & 0.86 & 0.00 & 0.00 \\ 
    CIFAR100 & DINO v1 ViT-B/16 & Standard & 0.76 & 0.00 & 0.00 \\ \hline
    CIFAR100 & DINO v1 ViT-B/16 & DeACL & 0.72 & 0.55 & 0.03 \\ \midrule
    
    STL10 & DINO v2 ViT-S/14 & Standard & 0.98 & 0.06 & 0.00 \\
    STL10 & DINO v2 ViT-B/14  & Standard & 0.99 & 0.20 & 0.00 \\ 
    STL10 & DINO v1 ViT-B/16 & Standard & 0.98 & 0.00 & 0.00 \\ \hline
    STL10 & DINO v1 ViT-B/16 & DeACL & 0.97 & 0.83 & 0.23 \\ 
    \bottomrule
    \end{tabular}
    \caption{\textbf{Evaluating the robustness of SSL encoders for linear classification.} 
    We observe that \EmbedAttack is effective against all \textit{Standard} encoders. Robustness against \EmbedAttack is improved with DeACL fine-tuning, however, the downstream \PGD attack maintains its effectiveness, with the possible exception of linear classification on STL10 dataset, a subset of ImageNet.
    }
    \label{tab:classification}
\end{table}
\addtolength{\tabcolsep}{\xintu}

\renewcommand\xintu{0.5pt}
\addtolength{\tabcolsep}{\xintu}
\begin{table}[h!]
\vspace{0.5cm}
    \tiny
    \centering
    \setlength{\tabcolsep}{3pt}
    \resizebox{\columnwidth}{!}{
    \begin{tabular}{ccc ccc}
    \toprule 
    Dataset & SSL & Encoder & \textit{Clean} & \EmbedAttack & \SegPGD \\
    & Framework & Type & mIoU$\uparrow$ & mIoU$\uparrow$ & mIoU$\uparrow$ \\ \midrule
    ADE20k & DINOv2 ViT-S/14 & Standard & 0.42 & 0.01 & 0.01 \\
    ADE20k & DINOv2 ViT-B/14 & Standard & 0.45 & 0.00 & 0.01 \\ 
    ADE20k & DINOv1 ViT-B/16 & Standard & 0.27 & 0.01 & 0.01 \\ \hline
    ADE20k & DINOv1 ViT-B/16 & DeACL & 0.24 & 0.14 & 0.01 \\ \midrule
    
    CityScapes & DINOv2 ViT-S/14 & Standard & 0.65 & 0.02 & 0.01 \\
    CityScapes & DINOv2 ViT-B/14 & Standard & 0.68 & 0.03 & 0.00 \\ 
    CityScapes & DINOv1 ViT-B/16 & Standard & 0.45 & 0.06 & 0.06 \\ \hline
    CityScapes & DINOv1 ViT-B/16 & DeACL & 0.36 & 0.31 & 0.03 \\ \midrule
    
    PASCAL VOC 2012 & DINOv2 ViT-S/14 & Standard & 0.83 & 0.00 & 0.01 \\
    PASCAL VOC 2012 & DINOv2 ViT-B/14 & Standard & 0.83 & 0.00 & 0.01 \\ 
    PASCAL VOC 2012 & DINOv1 ViT-B/16 & Standard & 0.56 & 0.06 & 0.00 \\ \hline
    PASCAL VOC 2012 & DINOv1 ViT-B/16 & DeACL & 0.51 & 0.30 & 0.02 \\
    \bottomrule
    \end{tabular}
    }
    \caption{\textbf{Evaluating the robustness of SSL encoders for semantic segmentation.} We note that \EmbedAttack succeeds across all datasets and Standard encoders, achieving mIoU close to 0, performing on par with \SegPGD. Fine-tuning with DeACL limits the effectiveness of \EmbedAttack, while \SegPGD remains successful.
    }
    \label{tab:segmentation}
\end{table}
\addtolength{\tabcolsep}{-\xintu}
\ \\  
\textbf{Semantic segmentation.} Similarly to classification, a single linear layer is trained on patch embeddings, to obtain a low-resolution logit map. Next, we interpolate the logits to obtain a logit map of a resolution matching the size of $x$. The minimized objective is a pixel-wise cross-entropy loss. We evaluate encoders on ADE20k~\citep{ADE1, ADE2}, CityScapes~\citep{Cordts2016cityScapes}, and PASCAL VOC 2012~\citep{Everingham10pascalVoc}, and report mean Intersection over Union (mIoU$\uparrow$) scores in \Cref{tab:segmentation}. \EmbedAttack proves to be a potent downstream task-agnostic method of obtaining adversarial examples for the segmentation task, achieving mIoU of 0 for all clean encoders across all datasets. Similarly to the linear classification task, we note that fine-tuning with DeACL improves robustness against \EmbedAttack, however, it fails to achieve significant improvements for the downstream attack \SegPGD.

\renewcommand\xintu{0.5pt}
\addtolength{\tabcolsep}{\xintu}
\begin{table}[h!]
\vspace{0.5cm}
    \centering
    \setlength{\tabcolsep}{3pt} 
    \tiny
    \begin{tabular}{cc cc cc cc}
    \toprule 
    SSL & Encoder & \multicolumn{1}{c}{\textit{Clean}} & \multicolumn{1}{c}{\EmbedAttack} & \multicolumn{1}{c}{\DepthPGD} \\
    Framework & Type & RMSE$\downarrow$ & RMSE$\downarrow$ & RMSE$\downarrow$ \\[0.0cm]  \midrule
    DINO v2 ViT-S/14 & Standard & 0.49 & 1.54 & 2.60 \\
    DINO v2 ViT-B/14 & Standard & 0.46 & 1.29 & 2.74 \\
    DINO v1 ViT-B/16 & Standard & 0.61& 1.28& 1.79\\
    \hline
    DINO v1 ViT-B/16 & DeACL & 0.68 & 0.92& 1.71 \\ %
    \bottomrule
    \end{tabular}
    \caption{\textbf{Evaluating the robustness of SSL encoders for depth estimation.} 
    On the NYU-Depth v2 dataset, we observe that both types of attacks remain effective against the DeACL defense.
    }
    \label{tab:depth_estimation}
\end{table}
\addtolength{\tabcolsep}{-\xintu}

\begin{figure*}[t!]
    \centering
    \begin{subfigure}[t]{0.33\linewidth}
        \centering
        \includegraphics[width=\textwidth, trim={0cm 0cm 0cm 0cm}]{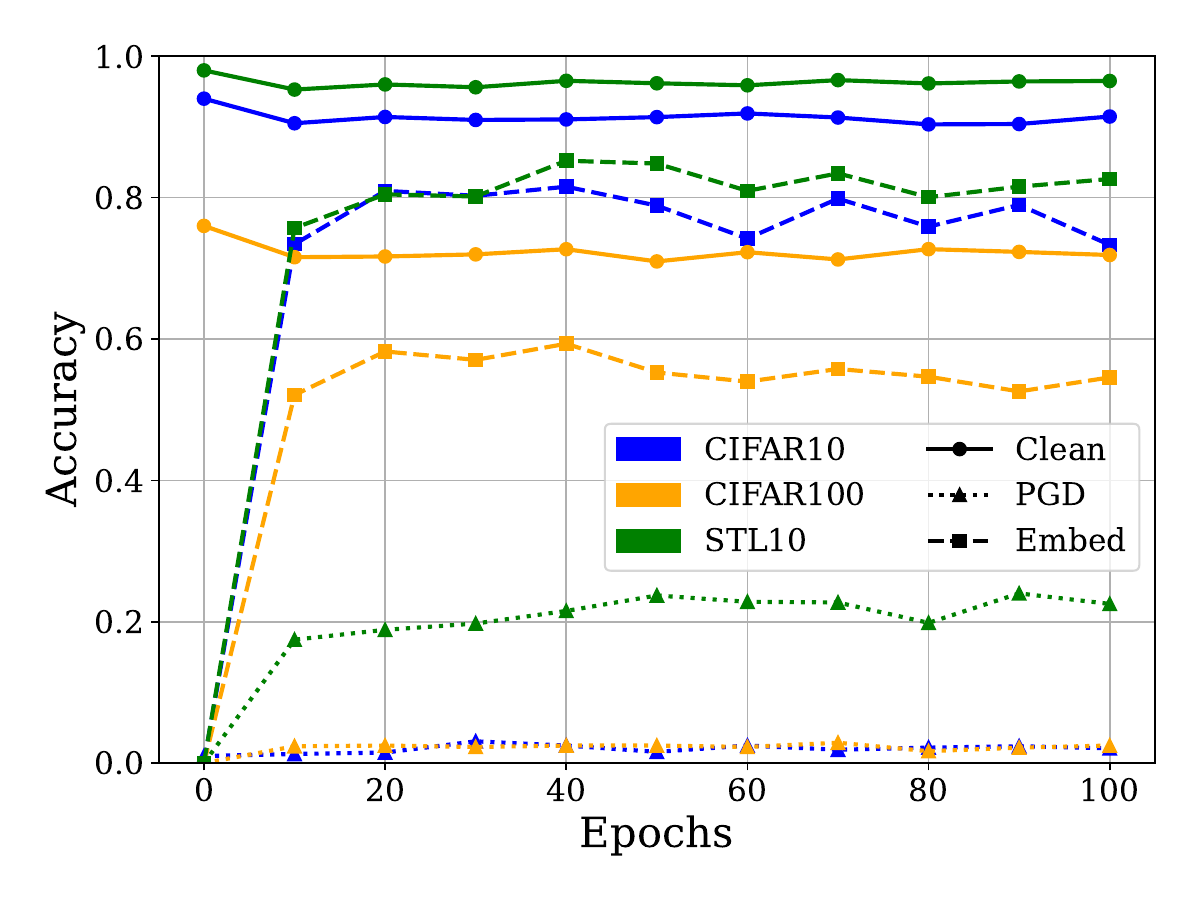}
        \caption{Classification}
        \label{fig:classification}
    \end{subfigure}
    \hfill
    \begin{subfigure}[t]{0.33\linewidth}
        \centering
        \includegraphics[width=\textwidth, trim={0cm 0cm 0cm 0cm}]{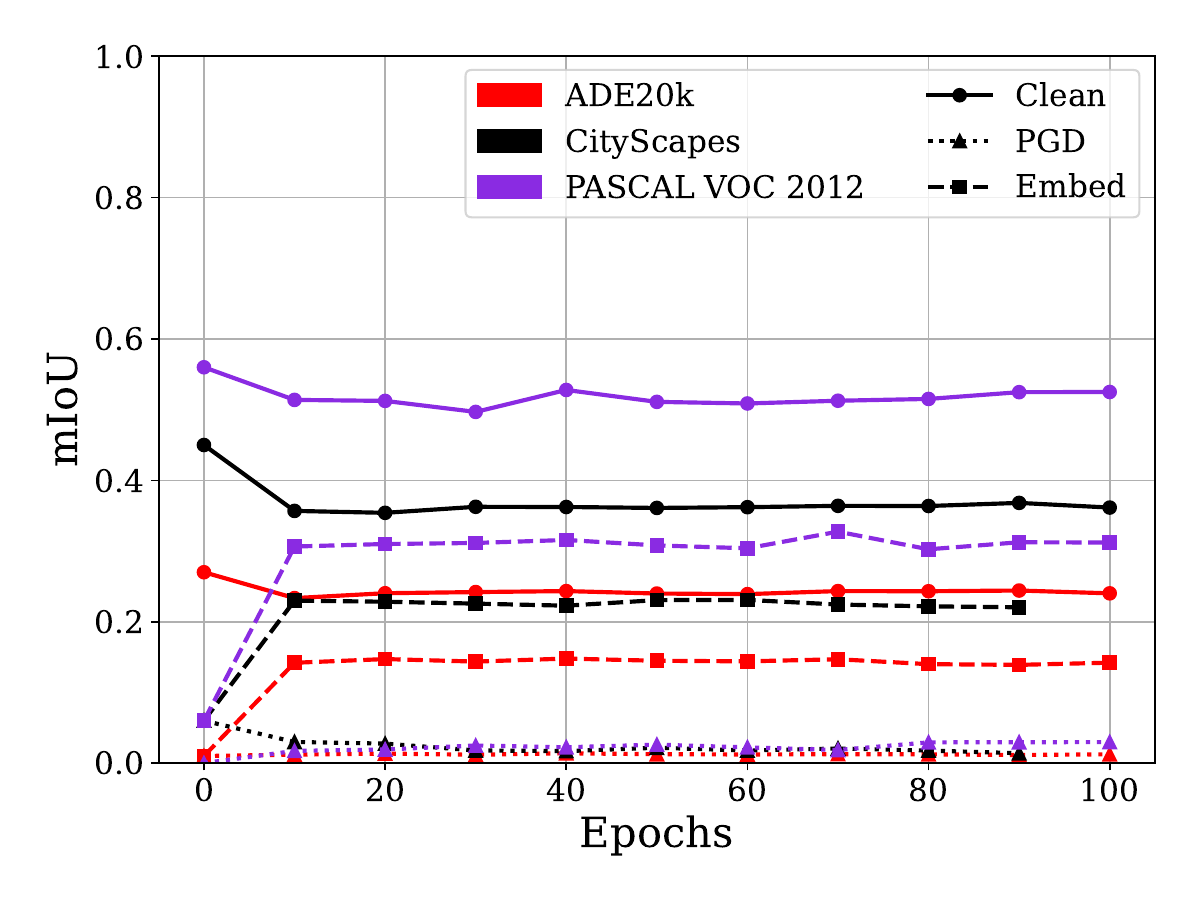}
        \caption{Segmentation}
        \label{fig:segmentation}
    \end{subfigure}
    \hfill
    \begin{subfigure}[t]{0.33\linewidth}
        \centering
        \includegraphics[width=\textwidth, trim={0cm 0cm 0cm 0cm}]{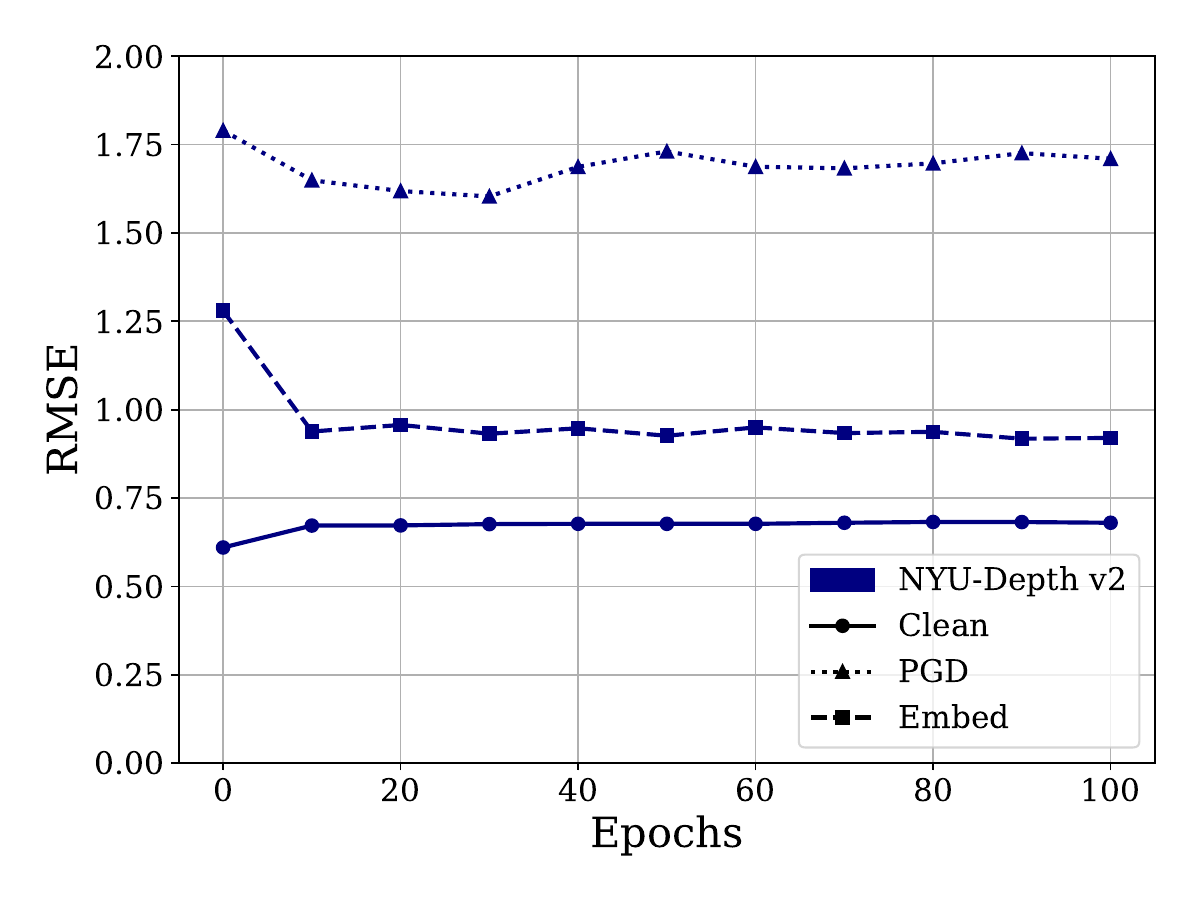}
        \caption{Depth estimation}
        \label{fig:depth_estimation}
    \end{subfigure}
    \caption{\textbf{Evolution of robustness on different downstream tasks during DeACL fine-tuning.} \Cref{fig:classification} presents classification accuracy on clean data for the \PGD and \EmbedAttack attacks. \Cref{fig:segmentation} shows segmentation mIoU on clean data for the \SegPGD and \EmbedAttack attacks. In
    \Cref{fig:depth_estimation} we present depth estiation RMSE on clean data for the \DepthPGD and \EmbedAttack attacks. Line colors indicate different datasets and line styles indicate different reported metrics.}

    \label{fig:evalution_decl}
\end{figure*}

\textbf{Depth estimation.}
For depth estimation, following \hbox{\citet{oquab2024dinov2}}, we extract the final layer of the frozen transformer and concatenate the CLS token with each patch token. Then we apply bilinear upsampling to the tokens to enhance the resolution. Finally, we train a linear layer on top to estimate the depth of each pixel. We evaluate quality of the depth estimation using the standard Root Mean Square Error (RMSE) metric on the NYU-Depth-v2 dataset \cite{Silberman2012NYU}. Our results in the \Cref{tab:depth_estimation} show that the \EmbedAttack and \DepthPGD attacks significantly increase the RMSE. The only instance where the RMSE remains below 1 after an attack is with DeACL fine-tuning against the \EmbedAttack; however, this fine-tuning fails to provide a notable improvement in robustness against the \DepthPGD attack, similarly to the classification and semantic segmentation tasks.

\textbf{Evolution of Robustness During DeACL Fine-Tuning.}
\Cref{fig:evalution_decl} presents the dynamics of model robustness for different downstream tasks during DeACL fine-tuning.
Notably, robustness against \PGD-based attacks exhibits minimal improvement, remaining unchanged during this process. 
The only exception is the improvement in robustness of linear classification on the STL10 dataset (which is a subset of ImageNet) observed during the first 20 epochs of training.
We also observe that a relatively short period of fine-tuning---around 10 epochs---leads to noticeable improvements in robustness against \EmbedAttack. However, further fine-tuning iterations show diminishing returns, with robustness metrics plateauing.
Performance on clean data remains relatively stable throughout fine-tuning after a drop during the first 10 epochs. 
The simultaneous increase in robustness against \EmbedAttack and decrease in performance on clean data observed at the start of the fine-tuning process confirms the trade-off between clean and adversarial model performance. The observed dynamics hold true across all downstream tasks. Our findings indicate that the adversarial fine-tuning method proposed by \citet{zhang2022decoupled} exerts its greatest impact during the initial epochs, with little to no benefit from prolonging training to a larger (\textit{e.g.} 100) number of epochs. 

\section{Discussion and Conclusions}
SSL encoders are foundation models leveraged for a myriad of downstream vision tasks in critical domains, like autonomous driving~\cite{liu2021sg} or medical imaging~\cite{jiang2018medical}.  This motivates the necessity of ensuring the encoders' robustness. In this work, we argue that prior work on SSL encoder robustness mainly evaluates downstream classification tasks while leaving other popular tasks, such as semantic segmentation or depth estimation under-explored. Through our experimentation, we show that encoders are highly vulnerable to adversarial attacks on  multiple downstream tasks, which pose a significant risk. Our results also highlight that the defenses that were developed with downstream classification in mind also harm the downstream performance on classification and other tasks. This suggests that more fundamental work is required to make foundational SSL encoders robust and effective for a wide variety of tasks.

\textbf{Future directions for improving robustness.}
We observe that defenses against adversarial examples in SSL are effective only for a single attack type, namely \EmbedAttack. However, they remain ineffective for other perturbations, especially task-specific attacks like \PGD, \SegPGD, and \DepthPGD. To train SSL models that are simultaneously robust to multiple perturbation types, a potential solution is to apply multi-perturbation adversarial training, similar to the approach used for enhancing robustness in supervised models against various perturbations~\citep{tramer2019adversarial}, which involved concurrent adversarial training with first-order $\ell_{1}$, $\ell_{2}$, and $\ell_{\infty}$ attacks. Therefore, to enhance the robustness of SSL encoders, we should not only fine-tune them on adversarial examples in the embedding space but also potentially perform robust tuning for each intended downstream task.

\newpage

\bibliographystyle{plainnat}
\bibliography{main}

\newpage
\appendix
\onecolumn
\appendix
\section{Societal Impact}
Prior work on SSL encoder robustness has primarily focused on classification tasks, leading to a false sense of security among users. Our findings reveal that encoders are also susceptible to attack on other downstream tasks, underscoring the need for more comprehensive defenses. This paves the way for the development of robust solutions, thereby enhancing the trustworthiness and reliability of foundational SSL encoders for broader societal applications.

\section{Extended Related Work}
\label{sec:extended-related-work}

\subsection{Adversarial Robustness in SSL}
\label{app:related_work_robustness}
For supervised tasks, adversarial attacks produce imperceptible changes $\delta$ to an input $x$ that result in the model predicting an incorrect label $y$ \citep{biggio2013evasion, szegedy2014intriguing}. To increase robustness, adversarial training \citep{goodfellow2015explaining} incorporates the perturbed data with the correct label into the training data. Since SSL operates without labels, this approach is not directly applicable. The initial method towards robust SSL proposed by \citet{NaseerPurifier2020CVPR} introduces a \textit{purifier network} to defend against adversarial examples, which attempts to recover the original input from an adversarially perturbed version before inputting it to the encoder. 
\textbf{Robust contrastive learning (RoCL)}~\citep{kim2020adversarial} instead aims to make the encoder itself robust by maximizing the similarity between a random augmentation of a data point and its instance-wise adversarial perturbation. RoCL translates instance-level robustness to class-level robustness, at the cost of substantial degradation in clean performance.

\citet{ho2020CLAE} propose adversarial examples specifically designed to challenge contrastive learning methods. Using these adversarial examples, they develop a novel adversarial training algorithm for self-supervised learning, which they call Contrastive Learning with Adversarial Examples (CLAE). Compared to standard contrastive learning, CLAE creates more difficult positive pairs by using adversarial examples. Additionally, by optimizing over all images in a batch, CLAE produces more challenging negative pairs through adversarial training. In essence, CLAE strengthens contrastive learning models by exposing them to tailored adversarial attacks during training.

\citet{jiang2020robust} introduce \textbf{adversarial contrastive learning (ACL)} to improve robustness-aware self-supervised pre-training by learning representations that are consistent under \textit{both data augmentations and adversarial perturbations}. They extend SimCLR~\citep{chen2020simple} to learn robust representations by maximizing feature consistency between differently augmented views.
\citet{fan2021does} build on top of ACL and propose \textbf{AdvCL}, which leverages labels in addition to instance-level robustness to further boost robust performance. 
\citet{luo2023rethinking} propose \textbf{Dynamic Adversarial Contrastive Learning (DYNACL)} as an extension that uses pseudo-labels directly generated by the pre-trained encoder. All these methods require retraining the large SSL encoders from scratch to improve robustness which is highly impractical and computationally expensive. To solve the problem, \cite{zhang2022decoupled} propose a two-stage framework called \textbf{Decoupled Adversarial Contrastive Learning (DeACL)} which fine-tunes existing encoders for increased robustness. Therefore, the knowledge of a pre-trained encoder is distilled to a robust one. The objective for the distillation are to: (1) match the distilled encoder representations to those of the pre-trained encoder, and (2) bring the distilled encoder's representations of adversarial examples close to their clean counterparts. \textit{Closeness} is defined by cosine similarity, and adversarial examples are just those examples generated on the original trained encoder to maximize the distance to the original samples. A compelling aspect is that the decoupling approach of DeACL is not limited to contrastive learning - the original encoder could potentially leverage other self-supervised learning (SSL) methods. Only the distillation loss may need adaptation for SSL frameworks like MAE~\citep{mae}, where cosine similarity may not be optimal.
Through this approach, DeACL sets a new state-of-the-art by effectively and efficiently improving encoder robustness. This is achieved by decoupling the SSL pre-training stage from the adversarial fine-tuning stage. The flexibility of DeACL leaves room for exploring different SSL methods in the first pre-training stage. Given the many advantages of DeACL demonstrated thus far, we focus our evaluation on this approach.

\section{Hyperparameters}
\label{sec:hyperparameters_app}

\subsection{Further Insights on Depth Estimation}
\label{sec:depth-hyper}
The multi-scale gradient matching loss \cite{MegaDepthLi18} encourages smoother transitions in depth predictions and penalizes differences in log-depth gradients across multiple scales:
\begin{equation}
L_{\text{grad}} = \frac{1}{n} \sum_k \sum_i \vert \nabla_x R^k_i \vert + \vert \nabla_y R^k_i \vert.
\end{equation}
The loss is computed at multiple scales where $R_i^k$ represents the value of the log-depth difference at position $i$ and scale $k$. $\nabla_x$ and $\nabla_y$ denote the gradients in the $x$ and $y$ directions, respectively.

The pixel-wise depth loss \cite{AdaBins} measures the difference between the predicted and ground truth depth values in a scale-invariant manner:
\begin{equation}
    L_{\text{pixel}} = \alpha \sqrt{\frac{1}{T} \sum_i g_i^2 - \frac{\rho}{T^2} \left( \sum_i g_i \right)}.
\end{equation}
Where $g_i = \log \tilde{d}_i - \log d_i$, with $\tilde{d}_i$ representing the predicted depth and $d_i$ the ground truth depth. The parameters $\alpha$ and $\rho$ are set to 1 and 0.85 in our experiments.

The final loss we use is $\frac{1}{2}L_{\text{grad}} + L_{\text{pixel}}$.

\subsection{DeACL fine-tuning}
\label{sec:hyperparameters_deacl_app}

In this section, we describe the hyperparameters we adopt to perform the adversarial fine-tuning proposed by \citep{zhang2022decoupled} on DINOv1 with ViT B/16 backbone. We use a learning rate of 0.05 with a \textit{cosine} scheduler and 10 epochs of warmup. We fine-tuned the model for 100 epochs with a SGD optimizer (momentum 0.9) and batch size of 128. The adversarial perturbation budget  $\varepsilon$ was set to 4/255. We did not use weight decay. We employed random crops, and random horizontal and vertical flips as training-time augmentations.

\end{document}